\documentclass[journal]{IEEEtran}
\usepackage{amsmath,graphicx,amsfonts,epsfig,epstopdf,datetime,algorithm,algorithmic,multirow,xcolor,subfig}

\begin{document}
%
\title{Deep Speaker Vectors for Semi Text-independent Speaker Verification}
%
%
%

\author{Lantian~Li,
        Dong~Wang~\IEEEmembership{Member,~IEEE},
        Zhiyong~Zhang,
        Thomas~Fang~Zheng$^*$~\IEEEmembership{Senior Member,~IEEE}
\thanks{This work was supported by the National Natural Science Foundation of China under Grant No. 61371136 and No. 61271389, it was also supported by the National Basic Research Program  (973 Program) of China under Grant No. 2013CB329302. The authors are with Division of Technical Innovation and Development of Tsinghua National Laboratory for Information Science and Technology and Research Institute of Information Technology (RIIT) of Tsinghua University. This paper is also supported by Sinovoice and Pachira. (Corresponding e-mail: fzheng@tsinghua.edu.cn)}}

\markboth{Journal of \LaTeX\ Class Files,~Vol.~13, No.~9, September~2014}%
{Shell \MakeLowercase{\textit{et al.}}: Bare Demo of IEEEtran.cls for Journals}

\maketitle

\begin{abstract}
Recent research shows that deep neural networks (DNNs) can be used to extract deep speaker vectors (d-vectors) that preserve speaker
characteristics and can be used in speaker verification. This new method has been tested on text-dependent speaker verification tasks,
and improvement was reported when combined with the conventional i-vector method.

This paper extends the d-vector approach to semi text-independent speaker verification tasks, i.e., the text of the speech is in a limited
set of short phrases. We explore various settings of the DNN structure used for d-vector extraction, and present a phone-dependent
training which employs the posterior features obtained from an ASR system. The experimental results show that it is possible to
apply d-vectors on semi text-independent speaker recognition, and the phone-dependent training improves system performance.

\end{abstract}


\IEEEpeerreviewmaketitle

\section{Introduction}

\IEEEPARstart{S}{peaker} verification, also known as voiceprint verification, is an important biometric authentication
technique that has been widely used to verify speakers' identities. According to the text that are
allowed to speak in enrollment and test, speaker verification systems can be categorized into either text-dependent or
text-independent. While a text-dependent system requires the same words/sentences to be spoken in
enrollment and test, a text-independent system permits any words to speak. This paper focuses on a semi text-independent
scenario where the words for enrollment and test are constrained in a limited set of short phrases, e.g.,
`turn on the radio'. With this limitation, people
can speak different sentences in enrollment and test while the system performance is
not significantly deteriorated, which makes the system more acceptable in practice.

Most of the successful approaches to speaker verification are based on generative models and
with unsupervised learning, e.g., the
famous Gaussian mixture model-universal background model (GMM-UBM) framework~\cite{ES2}.
A number of advanced models have been proposed based on the GMM-UBM architecture, among which
the i-vector model~\cite{ES3} ~\cite{ES4} is perhaps the most successful. Despite the impressive success,
the GMM-UBM model and the subsequent i-vector model share the
intrinsic disadvantage of all unsupervised learning methods: the goal of the model training is to
describe the distributions of the acoustic features, instead of discriminating speakers.

This problem can be solved in two directions. The first direction is to employ various discriminative models
to enhance the generative framework. For example, the SVM model for GMM-UBMs~\cite{ES5}, and the
PLDA model for i-vectors~\cite{ES6}. All these approaches provide significant improvement over the baseline.
Another direction is to look for more discriminative features, i.e., the features that
are more sensitive to speaker change and largely invariant to change of other irrelevant factors,
such as phone contents and channels~\cite{ES7}. However, the improvement obtained by the `feature engineering' is much less
significant compared to the achievements obtained by the discriminative models such as SVM and PLDA.
A possible reason is that most of the features are human-crafted and thus tend to be suboptimal in
practical usage.

Recent research on deep learning offers a new idea of `feature learning'. It has been
shown that with a deep neural network (DNN), task-oriented features can be learned
layer by layer from very raw features. For example in automatic speech recognition (ASR),
phone-discriminative features can be learned from spectrum or filter bank energies (Fbanks).
The learned features are very powerful and have defeated the conventional feature based on Mel frequency cepstral coefficients (MFCCs)
that has dominated in ASR for several decades~\cite{ES8}.

This favorable property of DNNs in learning task-oriented features can be utilized to
learn speaker-discriminative features as well. A recent study shows that this is possible
at least in text-dependent tasks~\cite{ES1}. The authors constructed a DNN model and
set the training objective as to discriminate a set of speakers, and for each frame,
the speaker-discriminative features were read from the activations of the last hidden layer.
They tested the method on a foot-print text-dependent speaker verification task
(only a short phrase `ok, google').
The experimental results showed that reasonable performance can
be achieved with the DNN-based features, although it is still difficult to compete the i-vector baseline.

In this paper, we extend the application of the DNN-based feature learning approach to semi text-independent tasks,
and present a phone-dependent training which involves phone posteriors obtained from an ASR system
in the training. The experimental results show that the DNN-based feature learning
works well on text-independent tasks, actually even better than on text-dependent tasks,
and the phone-dependent training offers marginal but consistent gains.

The rest of this paper is organized as follows. Section~\ref{sec:rel} describes the
related work, and Section~\ref{sec:theory} presents the DNN-based speaker feature learning.
The experiments are presented in Section~\ref{sec:exp}, and Section~\ref{sec:conl}
concludes the paper.

\section{Related work}
\label{sec:rel}

This paper follows the work in~\cite{ES1}. The difference is that we extend the application of the
DNN-based feature learning approach to semi text-independent tasks, and we introduce
a phone-dependent training. Due to the mismatched content of the enrollment and test
speech, our task is more challenging.

The DNN model has been employed in speaker verification in other ways. For example,
in~\cite{ES9}, DNNs trained for ASR were used to replace the UBM model to derive
the acoustic statistics for i-vector model training. In~\cite{ES10}, a DNN was used to
replace PLDA to improve discriminative capability of i-vectors. All these methods
rely on the generative framework, i.e., the i-vector model. The DNN-based
feature learning presented in this paper is purely discriminative, without
any generative model involved.

\section{DNN-based feature learning}
\label{sec:theory}

This section presents the DNN-based feature learning. We first describe the main
structure of the model and the learning process, and propose the phone-dependent learning.
Finally the difference between the i-vector approach and the DNN-based approach
is discussed.

\subsection{DNN-based feature extraction}

It is well-known that DNNs can learn task-oriented features from raw features
layer by layer. This property has been employed in ASR where
phone-discriminative features are learned from very low-level
features such as Fbanks or even spectrum~\cite{ES8}. It has been shown
that with a well-trained DNN, variations irrelevant to the learning
task are gradually eliminated when the input feature
is propagated through the DNN structure layer by layer.
This feature learning is so powerful that in ASR, the primary Fbank feature has
defeated the MFCC feature that was carefully designed by people and has
dominated in ASR for several decades.

This property can be also employed to learn speaker-discriminative features. Actually
researchers have put much effort in looking for features that are more discriminative
for speakers~\cite{ES7}, but the effort is mostly vain and the MFCC is still
the most popular choice. The success
of DNNs in ASR suggests a new direction that speaker-discriminative
features can be learned from data instead of crafted by hand. The learning can be
easily done and the process is rather similar as in ASR, with the only difference
that in speaker verification, the learning goal
is to discriminate different speakers.


\begin{figure}[htb]
   \centering
   \includegraphics[width=5cm]{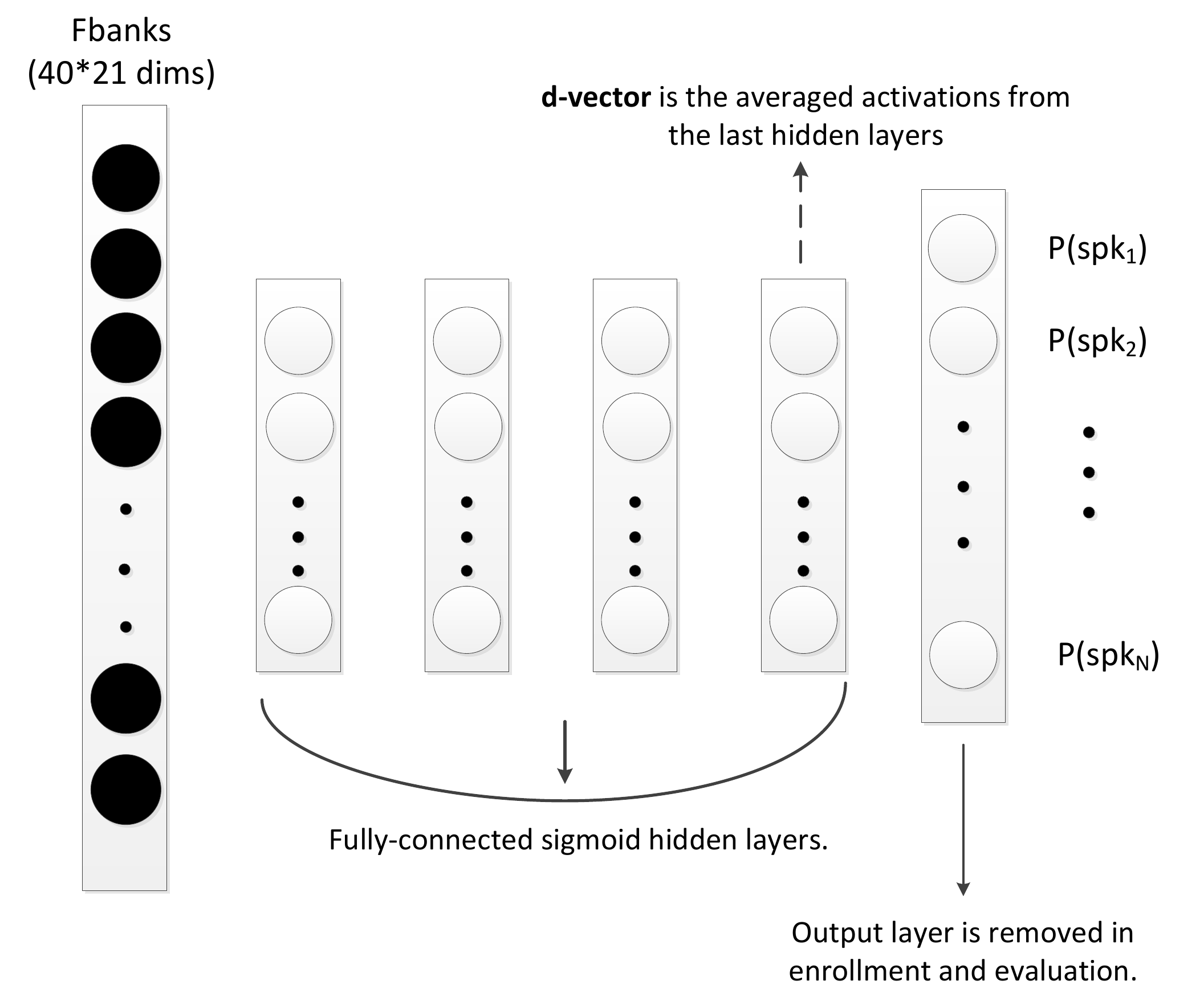}
   \caption{{\it The DNN structure used for learning speaker-discriminative features.}}
   \label{fig:1}
\end{figure}

Fig.~\ref{fig:1} presents the DNN structure used for the speaker-discriminative feature learning.
Following the convention of ASR, the input layer involves a window of 40-dimensional Fbanks.
In this work, the window size is set to $21$, which was found to be optimal in our work.
There are $4$ hidden layers, and each consists of $200$
units. The units of the output layer correspond to the speakers in the training data, and the number is $80$ in our experiment.
The 1-hot encoding scheme is used to label the target, and the training
criterion is set to cross entropy. The learning rate is set to $0.008$ at the beginning, and is
halved whenever no improvement on a cross-validation (CV) set is found. The training process stops when the learning
rate is too small and the improvement on the CV set is too marginal.

Once the DNN has been trained successfully, the speaker-discriminative features can be read from
any hidden layer. More the layer is close to the output, more the features are speaker-discriminative.
Our experiments show that features extracted from the last hidden layer perform the best, which is
similar to the observation in~\cite{ES1}.

In the test phase,  the features are extracted for all the frames of the given utterance, and the features are
averaged to form a speaker vector. Following the nomenclature in~\cite{ES1}, we call this speaker vector as
`d-vector'. Similar to i-vectors, a d-vector represents the speaker identity of an utterance in the speaker space.
The same methods used for i-vectors can be used for d-vectors to conduct the test, for example by computing the cosine distance or applying PLDA.

\subsection{Phone-dependent training}

A potential problem of the DNN-based speaker-discriminative feature learning described in the previous section
is that it is a `blind learning', i.e., the features are learned from raw data
without any prior information. This means that the learning purely relies on the complex deep structure
of the DNN model and a large amount of data to discover speaker-discriminative patterns. If the training data is
abundant, this is often not a problem; however in tasks with a limited amount of data,
for instance the semi text-independent task in our hand, this blind learning tends to be
difficult because there are too many speaker-irrelevant variations
involved in the raw data, particularly phone contents.

\begin{figure}[htb]
   \centering
   \includegraphics[width=5cm]{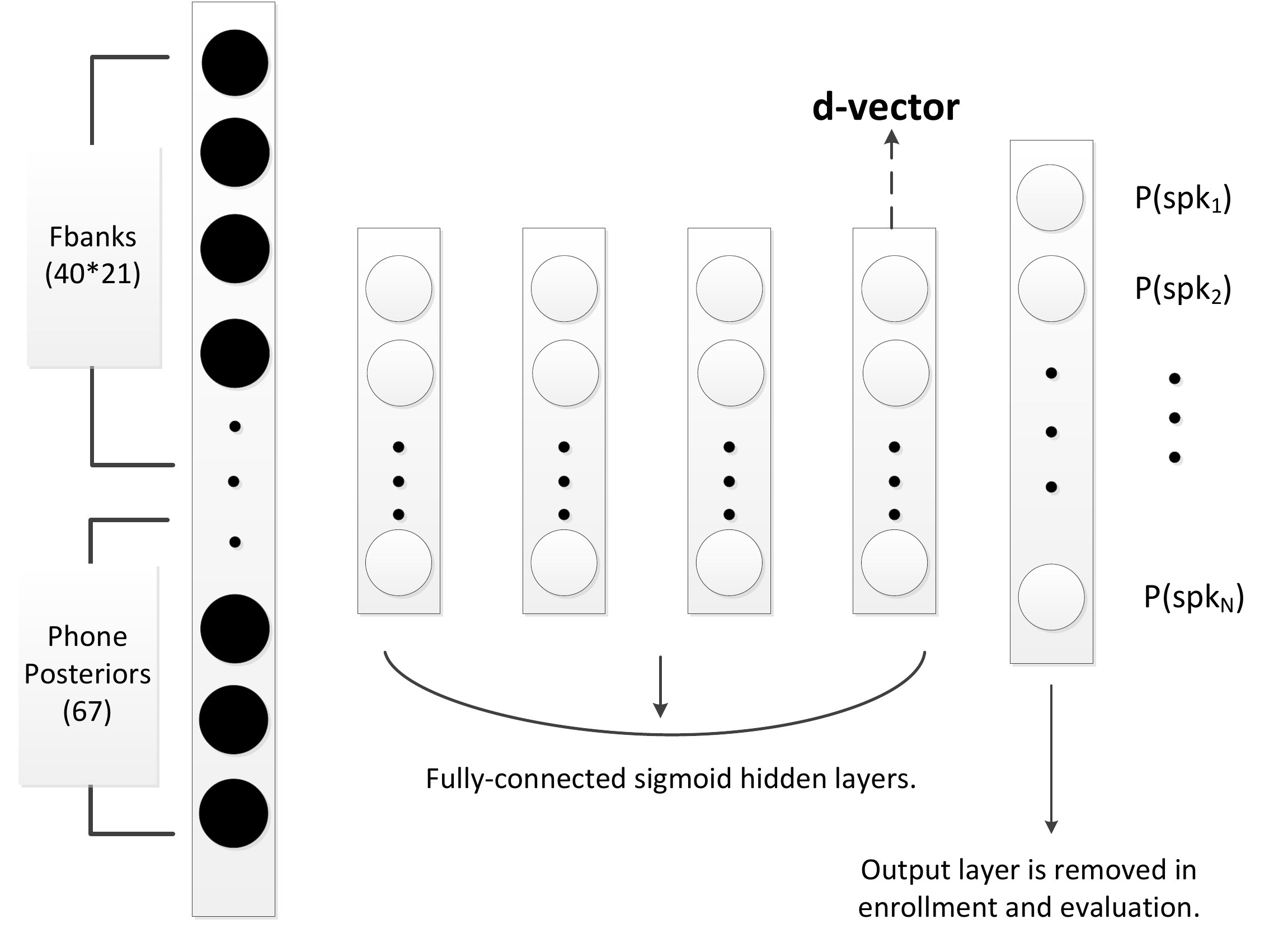}
   \caption{{\it The DNN structure used for phone-dependent training.}}
   \label{fig:phdnn}
\end{figure}

A possible solution is to inform the DNN which phone the current input frame belongs to. This can be simply achieved
by adding a phone indicator in the DNN input. However, it is often not easy to get the phone alignment
for the speech data. An alternative to the phone indicator is a vector of phone posterior probabilities, which
can be easily obtained from any phone discriminant model. In this work, we choose
a DNN model that was trained for an ASR system to produce the phone posteriors. Fig.~\ref{fig:phdnn} illustrates
the DNN structure with the phone posterior vector involved in the input. The training process for the new structure
does not change.

We note that this phone-dependent training is more important for text-independent recognition.
For the text-dependent recognition, the acoustic features are limited in a small set of phones,
and so involving the phone information in the training does not help much.


\subsection{Comparison between i-vectors and d-vectors}

The two kinds of speaker vectors, the d-vector and the i-vector, are fundamentally different.
I-vectors are based on a linear Gaussian model, for which the learning is unsupervised
and the learning criterion is maximum likelihood on acoustic features. In contrast, d-vectors are
based on neural networks, for which the learning is supervised, and the learning criterion is maximum
discrimination for speakers. This difference in model structures and learning methods leads to
significant different properties of these two vectors.

First, the i-vector is `descriptive', which represents the speaker
by constructing a GMM (derived from the i-vector) to fit the acoustic features.
In contrast, the d-vector is `discriminative', which represents
the speaker by removing speaker-irrelevant variance.

Second, the i-vector can be regarded as a `global' speaker description, which is
inferred from `all' the frames of an utterance; however the d-vector is a `local' description,
which is inferred from `each' frame, and only the context information is used in the inference. This
means that the d-vector tends to be more superior with a short utterance, while
the i-vector tends to perform better with a relative long utterance.

Third, the i-vector approach more relies on the enrollment data to form a reasonable
distribution that can be used to discriminate different speakers;
whereas the d-vector approach more relies on the `universal' data
to learn speaker-discriminative features. This means that a large amount
of training data (labelled with speakers) is much more important and useful for
the d-vector approach.

\section{Experiments}
\label{sec:exp}

\subsection{Database}

The experiments are performed on a short phrase speech database provided by Pachira. The entire
database contains recordings of $10$ short phrases from $100$ speakers (gender balanced), and each phrase
contains $2\sim5$ Chinese characters. For each speaker, every phrase is recorded $15$ times,
amounting to $150$ utterances per speaker.

The training set involves $80$ randomly selected speakers, which results in $12000$ utterances in total.
To prevent over-fitting, a cross-validation (CV) set containing $1000$ utterances is selected
from the training data, and the remaining $11000$ utterances are used for model training, including the
DNN model in the d-vector approach, and the UBM, the T matrix, the LDA and PLDA model in the i-vector approach.

The evaluation set consists of the remaining $20$ speakers. In the text-dependent experiment, the evaluation is
performed for each particular phrase; and in the semi text-independent experiment, all the utterances in the
evaluation set ($3000$ in total) are cross evaluated, resulting in $223500$ target trials
and $4275000$ non-target trials.

\subsection{Text-dependent recognition}

The first experiment investigates the performance of the d-vector approach on text-dependent speaker verification tasks, and compare
it to the i-vector baseline. A similar work has been reported in~\cite{ES1}, here we just reproduce that work
and propose some improvement by leveraging text-independent data.

For clearance, we report the results on two randomly selected phrases, denoted by P1 and P2 respectively. For each
phrase, the corresponding utterances are selected from the training set to train the i-vector system and the d-vector
system respectively, and the corresponding utterances in the evaluation set are selected to perform the test.
This means that the training data for each phrase consists of $1200$ utterances, and the test consists of $300$ utterances.
For the i-vector system, the number of Gaussian mixtures of the UBM is $64$, and the i-vector dimension is $200$. These
values have been chosen to optimize the performance. The DNN architecture for the d-vector system has been shown
in Section~\ref{sec:theory}. For a fair comparison, the dimension of the d-vector is set to $200$ as well.

The tests are based on three scoring methods: the basic cosine distance,
the cosine distance after reducing the dimension to $80$ by LDA, and the score provided by PLDA. Table~\ref{tab:text-d}
reports the  results in terms of
equal error rate (EER). It can be seen that the d-vector system obtains reasonable performance, however, the results are
much worse than those with the i-vector system. Similar observations have been reported in~\cite{ES1}.

\begin{table}[th]
        \centering
          \caption{EER results on text-dependent task}
          \label{tab:text-d}
          \begin{tabular}{l|c|c|c|c}
            \hline
                     & &\multicolumn{3}{|c}{EER\%} \\
                     \hline
                     &Phrase & Cosine & LDA & PLDA \\
            \hline
                i-vector & P1 & 4.91 & 4.62 & 4.05 \\
                d-vector & P1 & 12.05  & 9.52 & 10.76 \\
            \hline
                i-vector & P2 & 3.86 & 3.10 & 2.76 \\
                d-vector & P2 & 8.86 & 7.00 & 8.90 \\
            \hline
          \end{tabular}
\end{table}

As discussed in Section~\ref{sec:theory}, the DNN model of the d-vector system can be enhanced by borrowing
data from text-independent tasks. The results are reported in Table~\ref{tab:dnn-extra}.
It can be observed that with more training data, the performance of d-vector systems is generally improved, despite
that the extra data are recordings of other phrases. Another observation is that with more training data,
the PLDA model tends to be less effective. This can be possibly explained by the fact that d-vectors
are derived from activations of neural network units and so probably do not fit the linear Gaussian model that PLDA assumes.

\begin{table}[th]
        \centering
          \caption{EER results with additional data}
          \label{tab:dnn-extra}
          \begin{tabular}{l|l|c|c|c}
            \hline
                     & &\multicolumn{3}{|c}{EER\%} \\
                     \hline
                Phrase     & Training  & Cosine & LDA & PLDA \\
           \hline
                P1 &P1 & 12.05  & 9.52 & 10.76 \\
                P1 &P1,P2 & 11.57 & 8.29 & 10.57 \\
                P1 &P1,P2,...,P15& 11.14 & 8.14 & 11.00 \\
            \hline
                P2 &P2 & 8.86  & 7.00 & 8.90 \\
                P2 &P1,P2 & 7.95 & 5.81 & 6.91 \\
                P2 &P1,P2,...,P15 & 8.33 & 5.43 & 7.95 \\
            \hline
          \end{tabular}
\end{table}

\subsection{Semi text-independent recognition}

This experiment examines the d-vector approach on the semi text-independent task.
The dimension of both i-vectors and d-vectors is fixed to $200$, and the dimension
of the LDA-projected vectors is $80$.
In order to have the two systems involve the same amount of parameters,
the number of Gaussian components of the i-vector system is set to $128$.
All the utterances in the training dataset are used to train the DNN model and the i-vector model.

The results of the two systems are reported in Table~\ref{tab:txt-ind-ph}. It can be observed that with the simple cosine distance,
the d-vector system outperforms the i-vector system in a significant way. This demonstrates that the discriminatively
learned d-vectors are more discriminative for speakers when compared with the generatively learned i-vectors.
However, when the discriminative normalization methods (LDA and PLDA) are employed, the performance of the i-vector system
is significantly improved and better than that of the d-vector system.
The discriminative methods contribute very little to the d-vector system.
This is not supervising, as the d-vectors have been discriminative already.

\begin{figure}[t]
        \centering
        \includegraphics[width=5cm]{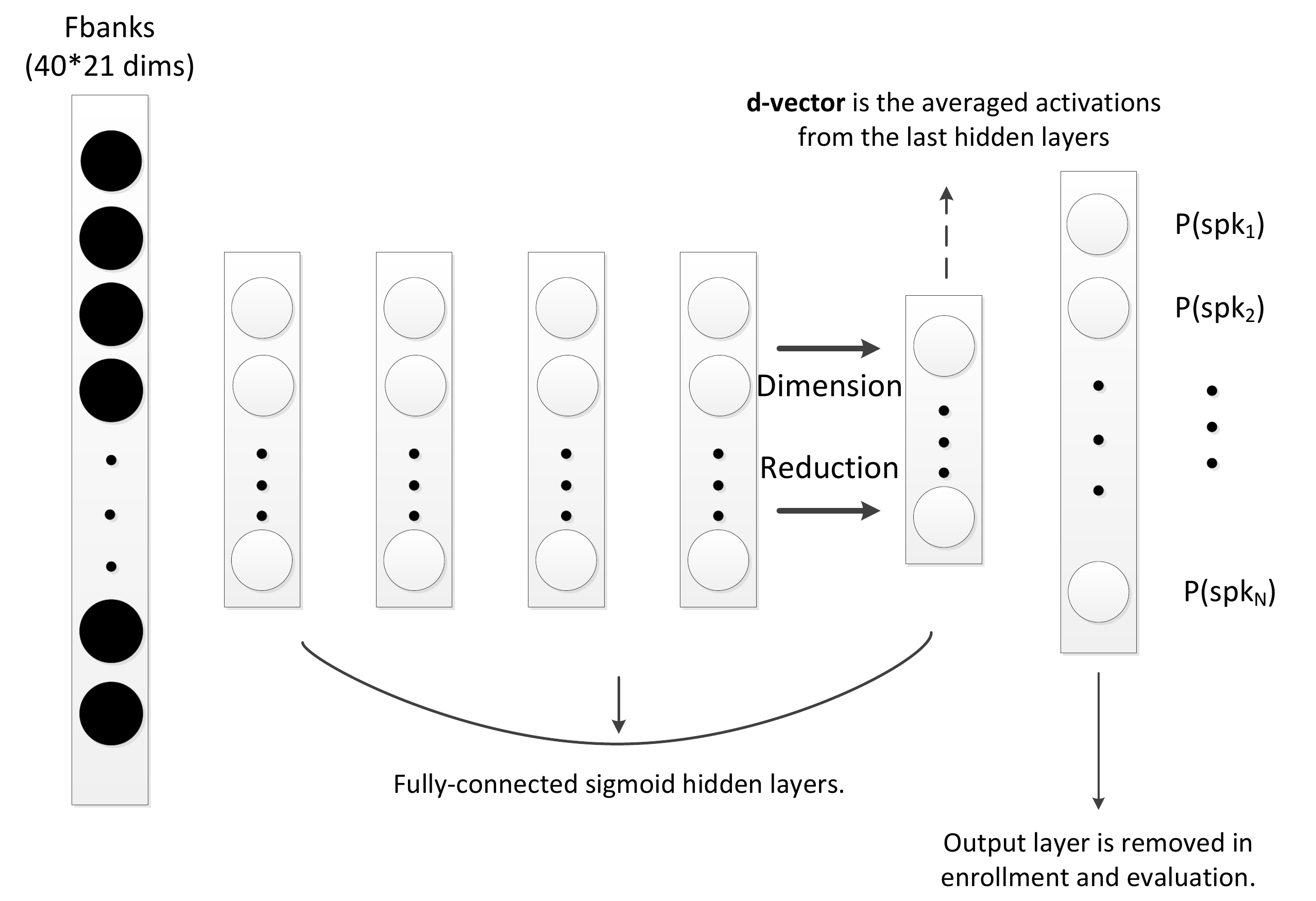}
        \caption{{\it The DNN structure with non-linear dimension reduction.}}
        \label{fig:nldr}
    \end{figure}

Nevertheless, the slight
improvement with LDA suggests that there is some redundancy in d-vectors.
Motivated by this idea, a new hidden layer with a small number of units is inserted into the DNN structure,
as shown in Fig.~\ref{fig:nldr}.  The dimension of the new layer is set to $100$, which is the best choice in our test. Compared
to LDA, this approach can be regarded as a non-linear dimension reduction (NLDR).
Additional performance is achieved with this method, as has been shown in the last column of Table~\ref{tab:txt-ind-ph}.

\subsection{Phone-dependent training}

In this experiment, the phone posteriors are included in the input of the DNN structure, as shown in Fig.~\ref{fig:phdnn}.
The phone posteriors are produced by a DNN model that was trained for ASR with a Chinese database consisting of $6000$
hours of speech data. The phone set consists of $66$ toneless initial and finals in Chinese, plus the silence phone.
The results are shown in the third row of Table~\ref{tab:txt-ind-ph}. It can be seen that the phone-dependent training leads to
marginal but consistent performance improvement for the d-vector system. The NLDR approach is also applied,
and an additional gain is obtained.

\begin{table}[th]
        \centering
          \caption{EER results on semi text-independent task}
          \label{tab:txt-ind-ph}

          \begin{tabular}{l|c|c|c|c|c}
            \hline
                    & PDTR   & cosine & LDA & PLDA    & NLDR\\
            \hline
            i-vector& -  & 19.32  & 11.09&  8.70  & - \\
            \hline
            d-vector& -  & 13.58  & 13.07 & 15.45 & 12.79 \\
            d-vector& +  & 13.21  & 12.76 & 15.48 & 12.55 \\
            \hline
          \end{tabular}
\end{table}

\subsection{Combination system}

\begin{figure}[t]
        \centering
        \includegraphics[width=5cm]{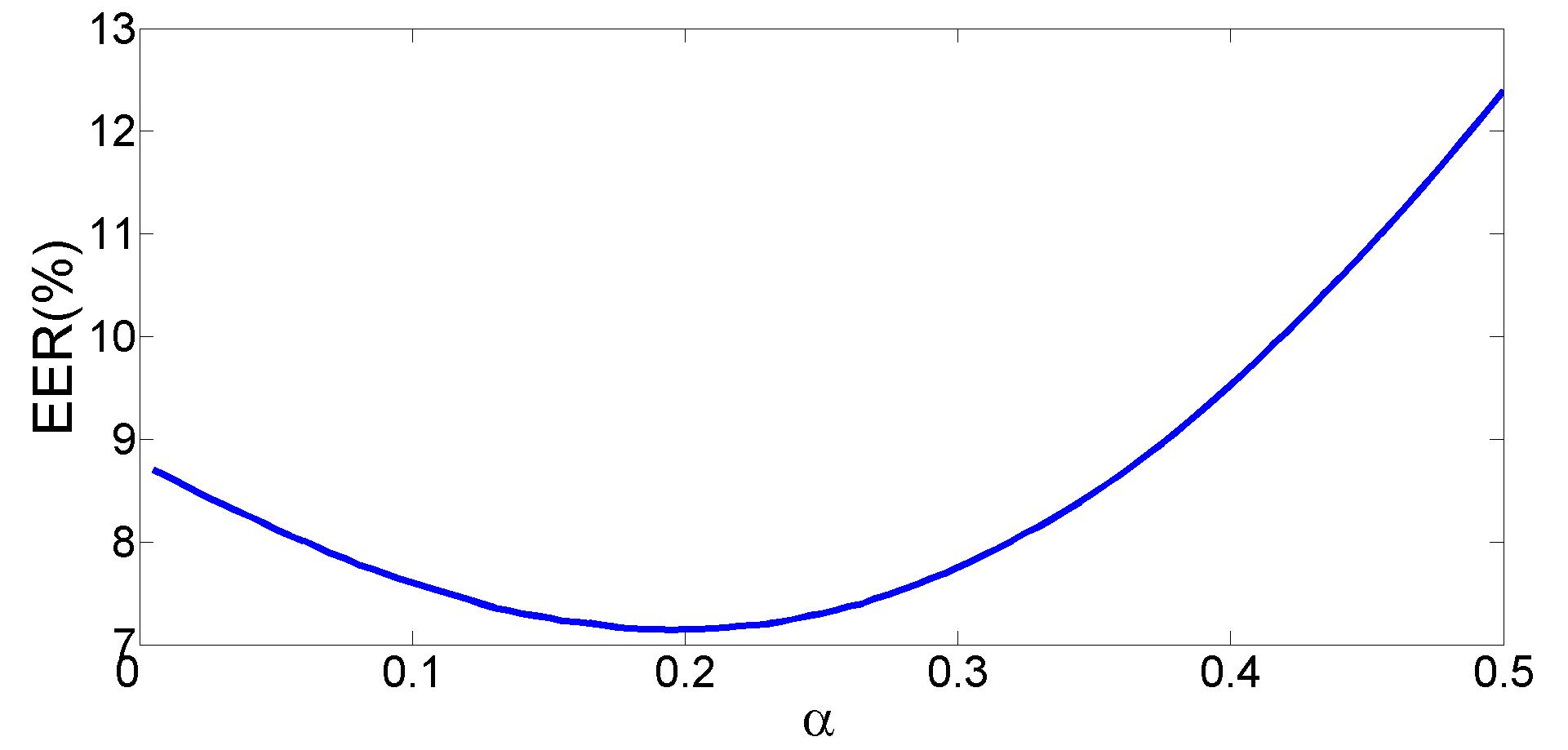}
        \caption{{\it The EER results of the d-vector and i-vector combination system. The x-axis represents the interpolation
        weight $\alpha$.}}
        \label{fig:comb}
    \end{figure}

Following~\cite{ES1}, we combine the best i-vector system (PLDA) and the best d-vector system (NLDR with phone-dependent training).
The combination is simply done by interpolating the scores obtained from the two systems.
The EER results with various values of the interpolation factor (denoted by $\alpha$) are drawn in Fig.~\ref{fig:comb}.
It can be seen that the combination leads to the better performance with an appropriate set of $\alpha$. The best
EER is $7.14\%$, which is the lowest EER we can obtain with this dataset so far.

\section{Conclusions}
\label{sec:conl}

This paper investigated DNN-based discriminative feature leaning for speaker recognition, and studied the performance of this approach on
a semi text-independent task. The experimental results demonstrated that the DNN-based approach can offer reasonable performance, and
outperformed the i-vector baseline with simple cosine distance. However, when discriminative normalization methods such as
LDA and PLDA are applied, the i-vector approach exhibits better performance.

Although it has not beat the i-vector approach at present, the d-vector approach is very promising and potentially can be improved in several
ways. Particularly, a powerful probabilistic model on d-vectors would deal with inter-frame uncertainty and so may considerably enhance system performance. We leave this as the future work.



\ifCLASSOPTIONcaptionsoff
  \newpage
\fi

\newpage



%

\bibliographystyle{IEEEtran}
\bibliography{dvector}

\end{document}